\documentclass{article}
\usepackage{lmodern}
\usepackage{fullpage}
\usepackage{enumitem}
\usepackage{subfig}
\usepackage{color}

\usepackage{algorithm}
\usepackage{algpseudocode}
\usepackage{amsmath}
\usepackage{amsthm}
\usepackage{stackengine}
\usepackage{epsfig}
\usepackage{multirow}
\usepackage{amsmath,amssymb}
\usepackage{bm}
\usepackage{color}
\usepackage{mdframed}
\usepackage{hyperref}
\hypersetup{breaklinks=true}
\usepackage{footnote}
\usepackage{natbib}
\usepackage[frozencache]{minted}

\def\x{{\mathbf{x}}}
\def\y{{\mathbf{y}}}
\def\w{{\mathbf{w}}}
\def\W{{\mathbf{W}}}
\def\b{{\mathbf{b}}}
\def\haty{{\hat{y}}}
\def\wb{{\mathbf{w}}}
\def\Real{{\mathbb{R}}}
\DeclareMathOperator*{\argmax}{arg\,max}

\title{Cyanure: An Open-Source Toolbox for Empirical Risk Minimization 
for Python, C++, and soon more.
}
\author{
Julien Mairal \\
   {Univ. Grenoble Alpes, Inria, CNRS, Grenoble INP, LJK, Grenoble, 38000}\\
         \texttt{julien.mairal@inria.fr}}

\begin{document}

\maketitle

\begin{abstract}
   Cyanure is an open-source C++ software package with a Python interface. 
   The goal of Cyanure is to provide state-of-the-art solvers for learning linear models,
   based on stochastic variance-reduced stochastic optimization with
   acceleration mechanisms.
   Cyanure can handle a large variety of loss functions (logistic, square,
   squared hinge, multinomial logistic) and regularization functions ($\ell_2$,
   $\ell_1$, elastic-net, fused Lasso, multi-task group Lasso).
   It provides a simple Python API, which is very close to that of scikit-learn,
   which should be extended to other languages such as R or Matlab in a near future.
\end{abstract}

\section{License and Citations}
Cyanure is distributed under BSD-3-Clause license. Even though this is non-legally binding, the author kindly ask users to cite the present arXiv document in their publications, as well as the publication related to the algorithm they have chosen (see Section~\ref{sec:formulations} for the related publications). If the default solver is used, the publication is most likely to be~\citep{lin2019inexact}.

\section{Main Features}
Cyanure is build upon several goals and principles:
\begin{itemize}
   \item {\textbf{Cyanure is memory efficient}}. If Cyanure accepts your dataset, it will never make a copy of it. Matrices can be provided in double or single precision. Sparse matrices (scipy/CSR format for Python, CSC for C++)
      can be provided with integers coded in 32 or 64-bits. When fitting an intercept, there is no need to add a column of 1's and there is no matrix copy as well. 
   \item {\textbf{Cyanure implements fast algorithms.}} Cyanure builds upon two algorithmic principles: (i) variance-reduced stochastic optimization; (ii) Nesterov of Quasi-Newton acceleration. Variance-reduced stochastic optimization algorithms are now popular, but tend to perform poorly when the objective function is badly conditioned. We observe large gains when combining these approaches with Quasi-Newton. 
   \item {\textbf{Cyanure only depends on your BLAS implementation.}} Cyanure does not depend on external libraries, except a BLAS library and numpy for Python. We show how to link with OpenBlas and Intel MKL in the python package, but any other BLAS implementation will do.
   \item {\textbf{Cyanure can handle many combinations of loss and regularization functions.}} Cyanure can handle a vast combination of loss functions (logistic, square, squared hinge, multiclass logistic) with regularization functions ($\ell_2$, $\ell_1$, elastic-net, fused lasso, multi-task group lasso).
   \item {\textbf{Cyanure provides optimization guarantees.}} We believe that reproducibility is important in research. For this reason, knowing if you have solved your problem when the algorithm stops is important. Cyanure provides such a guarantee with a mechanism called duality gap, see Appendix D.2 of~\citet{mairal2010sparse}.
   \item {\textbf{Cyanure is easy to use.}} We have developed a very simple API, relatively close to scikit-learn's API~\citep{scikit}, and provide also compatibility functions with scikit-learn in order to use Cyanure with minimum effort.
   \item {\textbf{Cyanure should not be only for Python.}} A python interface is provided for the C++ code, but it should be feasible to develop an interface for any language with a C++ API, such as R or Matlab. We are planning to develop such interfaces in the future.
\end{itemize}
Besides all these nice features, Cyanure has also probably some drawbacks, which we will let you discover by yourself.  

\section{Where to find Cyanure and how to install it?}
The webpage of the project is available at \url{http://julien.mairal.org/cyanure/} and detailed instructions are given there. The software package can be found on github and Pipy.

\section{Formulations}\label{sec:formulations}
Cyanure addresses the minimization of empirical risks, which covers a large
number of classical formulations such as logistic regression, support vector
machines with squared hinge loss (we do handle the regular hinge loss at the
moment), or multinomial logistic.

\paragraph{Univariate problems.} We consider a dataset of pairs of
labels/features $(y_i,\x_i)_{i=1,\ldots,n}$, where the features~$\x_i$ are
vectors in $\Real^p$ and the labels $y_i$ are in $\Real$ for regression
problems and in $\{-1,+1\}$ for classification.
Cyanure learns a predition function $h(\x)= \w^\top \x + b$ or $h(\x)=\text{sign}(\w^\top \x + b)$, respectively for regression and classification, 
where $\w$ in
$\Real^p$ represents the weights of a linear model and $b$ is an (optional)
intercept. Learning these parameters is achieved by minimizing
\begin{displaymath}
   \min_{\w \in \Real^p, b \in \Real} \frac{1}{n}\sum_{i=1}^n \ell(y_i, \wb^\top \x + b)  + \psi(\w), 
\end{displaymath}
where $\psi$ is a regularization function, which will be detailed later, and $\ell$ is a loss function that is chosen among the following choices
\begin{table}[h!]
   \centering
   \begin{tabular}{|c|c|c|}
      \hline
      Loss  & Label  &  $\ell(y, \hat{y})$ \\  
      \hline
      \texttt{square} &  $y \in \Real$ &  $\frac{1}{2}( y-\haty)^2$ \\ 
      \hline
      \texttt{logistic} &  $y \in \{-1,+1\}$ &  $ \log(1+e^{-y \hat{y}})$ \\ 
      \hline
      \texttt{sq-hinge} &  $y \in \{-1,+1\}$ &  $ \frac{1}{2}\max(0,1-y \hat{y})^2$ \\ 
      \hline
      \texttt{safe-logistic} &  $y \in \{-1,+1\}$ &  $e^{y \haty -1}- y \haty $~~~ if~~~ $ y \haty \leq 1$ and $0$ otherwise \\ 
      \hline
   \end{tabular}
   \caption{Loss functions used for univariate machine learning problems.} \label{table:loss}
\end{table}

\begin{figure}[h!]
   \centering
   \includegraphics[width=0.6\linewidth]{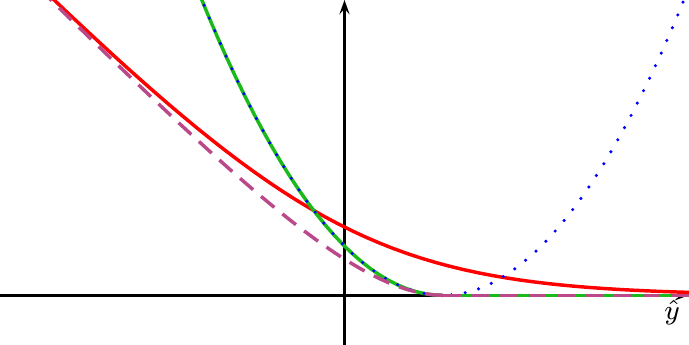}
   \caption{Four loss functions plotted for $y=1$; \texttt{square} loss is in dotted blue; \texttt{logistic} loss is in plain red; \texttt{sq-hinge} is in plain green; \texttt{safe-logistic} is in dashed magenta. All loss functions are smooth.}\label{fig:loss}
\end{figure}

\texttt{logistic} corresponds to the loss function used for logistic
regression. You may be used to a different (but equivalent) formula if you use
labels in $\{0,1\}$.  In order to illustrate the choice of loss functions, we
plot in Figure~\ref{fig:loss} their values as functions of $\haty$ when $y=+1$.
We note that all of them are differentiable---a requirement of our algorithms---,
which is why we do not handle the regular hinge loss.

These problems will be handled by the classes \texttt{BinaryClassifier} and
\texttt{Regression} for binary classification and regression problems,
respectively.  If you are used to scikit-learn~\citep{scikit}, we also provide the classes
\texttt{LinearSVC} and \texttt{LogisticRegression}, which are compatible with
scikit-learn's API (in particular, they use the parameter \texttt{C} instead of
$\lambda$).

The regularization function~$\psi$ is chosen among the following choices
\begin{table}[h!]
   \centering
   \begin{tabular}{|c|c|}
      \hline
      \multicolumn{2}{|c|}{Penalty $\psi$}     \\  
      \hline
      \texttt{l2} &   $\frac{\lambda}{2}\|\w\|_2^2$ \\ 
      \hline
      \texttt{l1} &   $\lambda\|\w\|_1$ \\ 
      \hline
      \texttt{elastic-net} &   $\lambda\|\w\|_1 + \frac{\lambda_2}{2}\|\w\|_2^2$ \\ 
      \hline
      \texttt{fused-lasso} &   $\lambda\sum_{j=2}^p|\w[j]-\w[j-1]| + \lambda_2\|\w\|_1 + \frac{\lambda_3}{2}\|\w\|_2^2$ \\ 
      \hline
      \texttt{none} &   $0$ \\ 
      \hline
      \hline
      \multicolumn{2}{{|c|}}{Constraint $\psi$}     \\  
      \hline
      \texttt{l1-ball} &  (constraint) $\|\wb\|_1 \leq \lambda$ \\ 
      \hline
      \texttt{l2-ball} &  (constraint) $\|\wb\|_2 \leq \lambda$ \\ 
      \hline
   \end{tabular}
   \caption{Regularization functions used for univariate machine learning problems.}\label{table:penalty}
\end{table}

Note that the intercept~$b$ is not regularized and the last two regularization
functions are encoding constraints on the variable~$\wb$.

\paragraph{Multivariate problems.}
We now consider a dataset of pairs of
labels/features $(\y_i,\x_i)_{i=1,\ldots,n}$, where the features~$\x_i$ are
vectors in $\Real^p$ and the labels $\y_i$ are in $\Real^k$ for multivariate regression
problems and in $\{1,2,\ldots,k\}$ for multiclass classifiation.
For regression, 
Cyanure learns a predition function $n(\x)= \W^\top \x + \b$, where $\W=[\w_1,\ldots,\w_k]$ in
$\Real^{p \times k}$ represents the weights of a linear model and $\b$ in $\Real^k$ is a set of (optional)
intercepts.
For classification, the prediction function is simply $h(\x)= \argmax_{j=1,\ldots,k} \w_j^\top \x + \b_j$.

Then, Cyanure optimizes the following cost function
\begin{displaymath}
   \min_{\W \in \Real^{p \times k}, \b \in \Real^k} \frac{1}{n}\sum_{i=1}^n \ell(\y_i, \W^\top \x + \b)  + \psi(\W). 
\end{displaymath}
The loss function may be chosen among the following ones
\begin{table}[h!]
   \centering
   \begin{tabular}{|p{3.5cm}|c|c|}
      \hline
      Loss  & Label  &  $\ell(\y, \hat{\y})$ \\  
      \hline
      \texttt{square} &  $\y \in \Real^k$ &  $\frac{1}{2}\|\y-\hat{\y}\|_2^2$ \\ 
      \hline
      \texttt{square},  
      \texttt{logistic},  
      \texttt{sq-hinge},
      \texttt{safe-logistic} &  $\{1,2,\ldots,k\} \Rightarrow \y \in \{-1,+1\}^k$  &  $\sum_{j=1}^k \tilde{\ell}(\y[j], \hat{\y}[j])$, with $\tilde{\ell}$ chosen from Table~\ref{table:loss}   \\ 
      \hline
      \texttt{multiclass-logistic} & $y \in \{1,2,\ldots,k\}$ &  $\sum_{j=1}^k \log\left(e^{\hat{\y}[j] - \hat{\y}[y]}  \right)  $  \\
      \hline
   \end{tabular}
   \caption{Loss functions used for multivariate machine learning problems. \texttt{multiclass-logistic} is also called the multinomial logistic loss function.}
\end{table}

The regularization functions may be chosen according to the following table.
\begin{table}[h!]
   \centering
   \begin{tabular}{|p{5cm}|c|}
      \hline
      \multicolumn{2}{|c|}{Penalty or constraint $\psi$}     \\  
      \hline
      \texttt{l2, l1, elastic-net, fused-lasso, none, l1-ball, l2-ball} & 
      $\sum_{j=1}^k \tilde{\psi}(\w_j)$, with $\tilde{\psi}$ chosen from Table~\ref{table:penalty} \\
      \hline
      \texttt{l1l2} & $\lambda \sum_{j=1}^p \| \W^j\|_2$ ($\W^j$ is the $j$-th row of $\W$) \\ 
      \hline
      \texttt{l1linf} & $\lambda \sum_{j=1}^p \| \W^j\|_\infty$  \\ 
      \hline
   \end{tabular}
   \caption{Regularization function for multivariate machine learning problems.}
\end{table}

In Cyanure, the classes \texttt{MultiVariateRegression} and \texttt{MultiClassifier} are devoted to multivariate machine learning problems.

\paragraph{Algorithms.}
Cyanure implements the following solvers:
\begin{itemize}
   \item \texttt{ista}: the basic proximal gradient descent method with line-search~\citep[see][]{fista};
   \item \texttt{fista}: its accelerated variant~\citep{fista};
   \item \texttt{qning-ista}: the Quasi-Newton variant of ISTA proposed by~\citet{lin2019inexact}; this is an effective variant of L-BFGS, which can handle composite optimization problems.
   \item \texttt{miso}: the MISO algorithm of~\citet{miso}, which may be seen as a primal variant of SDCA~\citep{sdca};
   \item \texttt{catalyst-miso}: its accelerated variant, by using the Catalyst approach of~\citet{lin2018catalyst};
   \item \texttt{qning-miso}: its Quasi-Newton variant, by using the QNing approach of~\citet{lin2019inexact};
   \item \texttt{svrg}: a non-cyclic variant of SVRG~\citep{proxsvrg}, see~\citep{kulunchakov2019estimate};
   \item \texttt{catalyst-svrg}: its accelerated variant by using Catalyst~\citep{lin2018catalyst};
   \item \texttt{qning-svrg}: its Quasi-Newton variant by using QNing~\citep{lin2019inexact};
   \item \texttt{acc-svrg}: a variant of SVRG with direct acceleration introduced in~\citep{kulunchakov2019estimate};
   \item \texttt{auto}: \texttt{auto} will use \texttt{qning-ista} for small problems, or \texttt{catalyst-miso}, or \texttt{qning-miso}, depending on the conditioning of the problem.
\end{itemize}

\section{How to use it?}
The online documentation provides information about the four main classes
\texttt{BinaryClassifier}, \texttt{Regression},
\texttt{MultiVariateRegression}, and \texttt{MultiClassifier}, and how to use
them. Below, we provide simple examples.
\paragraph{Examples for binary classification.}
The following code performs binary classification with $\ell_2$-regularized logistic regression, with no intercept, on the criteo dataset (21Gb, huge sparse matrix)
\begin{minted}{python}
import cyanure as cyan
#load criteo dataset 21Gb, n=45840617, p=999999
dataY=np.load('criteo_y.npz',allow_pickle=True); y=dataY['y']
X = scipy.sparse.load_npz('criteo_X.npz')
#normalize the rows of X in-place, without performing any copy
cyan.preprocess(X,normalize=True,columns=False) 
#declare a binary classifier for l2-logistic regression
classifier=cyan.BinaryClassifier(loss='logistic',penalty='l2')
# uses the auto solver by default, performs at most 500 epochs
classifier.fit(X,y,lambd=0.1/X.shape[0],nepochs=500,tol=1e-3,it0=5) 
\end{minted}

Before we comment the previous choices, let us 
run the above code on a regular three-years-old quad-core workstation with 32Gb of memory
(Intel Xeon CPU E5-1630 v4, 400\$ retail price). 
\begin{verbatim}
Matrix X, n=45840617, p=999999
*********************************
Catalyst Accelerator
MISO Solver
Incremental Solver with uniform sampling
Lipschitz constant: 0.250004
Logistic Loss is used
L2 regularization
Epoch: 5, primal objective: 0.456014, time: 92.5784
Best relative duality gap: 14383.9
Epoch: 10, primal objective: 0.450885, time: 227.593
Best relative duality gap: 1004.69
Epoch: 15, primal objective: 0.450728, time: 367.939
Best relative duality gap: 6.50049
Epoch: 20, primal objective: 0.450724, time: 502.954
Best relative duality gap: 0.068658
Epoch: 25, primal objective: 0.450724, time: 643.323
Best relative duality gap: 0.00173208
Epoch: 30, primal objective: 0.450724, time: 778.363
Best relative duality gap: 0.00173207
Epoch: 35, primal objective: 0.450724, time: 909.426
Best relative duality gap: 9.36947e-05
Time elapsed : 928.114
\end{verbatim}
The solver used was \texttt{catalyst-miso}; the problem was solved up to
accuracy $\varepsilon=0.001$ in about $15$mn after 35 epochs (without taking into account
the time to load the dataset from the hard drive). The regularization
parameter was chosen to be $\lambda=\frac{1}{10n}$, which is close to the
optimal one given by cross-validation.  Even though performing a grid search with
cross-validation would be more costly, it nevertheless shows that processing such 
a large dataset does not necessarily require to massively invest in Amazon EC2 credits,
GPUs, or distributed computing architectures.

In the next example, we use the squared hinge loss with
$\ell_1$-regularization, choosing the regularization parameter such that the
obtained solution has about $10\%$ non-zero coefficients.
We also fit an intercept. As shown below, the solution is obtained in 26s on a
laptop with a quad-core i7-8565U CPU.
\begin{minted}{python}
import cyanure as cyan
#load rcv1 dataset about 1Gb, n=781265, p=47152
data = np.load('rcv1.npz',allow_pickle=True); y=data['y']; X=data['X']
X = scipy.sparse.csc_matrix(X.all()).T # n x p matrix, csr format 
#normalize the rows of X in-place, without performing any copy
cyan.preprocess(X,normalize=True,columns=False) 
#declare a binary classifier for squared hinge loss + l1 regularization
classifier=cyan.BinaryClassifier(loss='sqhinge',penalty='l2')
# uses the auto solver by default, performs at most 500 epochs
classifier.fit(X,y,lambd=0.000005,nepochs=500,tol=1e-3) 
\end{minted}
which yields
\begin{verbatim}
Matrix X, n=781265, p=47152
Memory parameter: 20
*********************************
QNing Accelerator
MISO Solver
Incremental Solver with uniform sampling
Lipschitz constant: 1
Squared Hinge Loss is used
L1 regularization
Epoch: 10, primal objective: 0.0915524, time: 7.33038
Best relative duality gap: 0.387338
Epoch: 20, primal objective: 0.0915441, time: 15.524
Best relative duality gap: 0.00426003
Epoch: 30, primal objective: 0.0915441, time: 25.738
Best relative duality gap: 0.000312145
Time elapsed : 26.0225
Total additional line search steps: 8
Total skipping l-bfgs steps: 0
\end{verbatim}

\paragraph{Multiclass classification.}
Let us now do something a bit more involved and perform multinomial logistic regression on the
\texttt{ckn\_mnist} dataset (10 classes, $n=60\,000$, $p=2304$, dense matrix), with multi-task group lasso regularization,
using the same laptop as previously, and choosing a regularization parameter that yields a solution with $5\%$ non zero coefficients.
\begin{minted}{python}
import cyanure as cyan
#load ckn_mnist dataset 10 classes, n=60000, p=2304
data=np.load('ckn_mnist.npz'); y=data['y']; X=data['X']
#center and normalize the rows of X in-place, without performing any copy
cyan.preprocess(X,centering=True,normalize=True,columns=False) 
#declare a multinomial logistic classifier with group Lasso regularization
classifier=cyan.MultiClassifier(loss='multiclass-logistic',penalty='l1l2')
# uses the auto solver by default, performs at most 500 epochs
classifier.fit(X,y,lambd=0.0001,nepochs=500,tol=1e-3,it0=5) 
\end{minted}
\begin{verbatim}
Matrix X, n=60000, p=2304
Memory parameter: 20
*********************************
QNing Accelerator
MISO Solver
Incremental Solver with uniform sampling
Lipschitz constant: 0.25
Multiclass logistic Loss is used
Mixed L1-L2 norm regularization
Epoch: 5, primal objective: 0.340267, time: 30.2643
Best relative duality gap: 0.332051
Epoch: 10, primal objective: 0.337646, time: 62.0562
Best relative duality gap: 0.0695877
Epoch: 15, primal objective: 0.337337, time: 93.9541
Best relative duality gap: 0.0172626
Epoch: 20, primal objective: 0.337293, time: 125.683
Best relative duality gap: 0.0106066
Epoch: 25, primal objective: 0.337285, time: 170.044
Best relative duality gap: 0.00409663
Epoch: 30, primal objective: 0.337284, time: 214.419
Best relative duality gap: 0.000677961
Time elapsed : 215.074
Total additional line search steps: 4
Total skipping l-bfgs steps: 0
\end{verbatim}
Learning the multiclass classifier took about 3mn and 35s. To conclude, we provide a last more classical example
of learning l2-logistic regression classifiers on the same dataset, in a one-vs-all fashion.
\begin{minted}{python}
import cyanure as cyan
#load ckn_mnist dataset 10 classes, n=60000, p=2304
data=np.load('ckn_mnist.npz'); y=data['y']; X=data['X']
#center and normalize the rows of X in-place, without performing any copy
cyan.preprocess(X,centering=True,normalize=True,columns=False) 
#declare a multinomial logistic classifier with group Lasso regularization
classifier=cyan.MultiClassifier(loss='logistic',penalty='l2')
# uses the auto solver by default, performs at most 500 epochs
classifier.fit(X,y,lambd=0.01/X.shape[0],nepochs=500,tol=1e-3) 
\end{minted}
Then, the $10$ classifiers are learned in parallel using the four cpu cores
(still on the same laptop), which gives the following output after about $1$mn
\begin{verbatim}
Matrix X, n=60000, p=2304
Solver 4 has terminated after 30 epochs in 36.3953 seconds
   Primal objective: 0.00877348, relative duality gap: 8.54385e-05
Solver 8 has terminated after 30 epochs in 37.5156 seconds
   Primal objective: 0.0150244, relative duality gap: 0.000311491
Solver 9 has terminated after 30 epochs in 38.4993 seconds
   Primal objective: 0.0161167, relative duality gap: 0.000290268
Solver 7 has terminated after 30 epochs in 39.5971 seconds
   Primal objective: 0.0105672, relative duality gap: 6.49337e-05
Solver 0 has terminated after 40 epochs in 45.1612 seconds
   Primal objective: 0.00577768, relative duality gap: 3.6291e-05
Solver 6 has terminated after 40 epochs in 45.8909 seconds
   Primal objective: 0.00687928, relative duality gap: 0.000175357
Solver 2 has terminated after 40 epochs in 45.9899 seconds
   Primal objective: 0.0104324, relative duality gap: 1.63646e-06
Solver 5 has terminated after 40 epochs in 47.1608 seconds
   Primal objective: 0.00900643, relative duality gap: 3.42144e-05
Solver 3 has terminated after 30 epochs in 12.8874 seconds
   Primal objective: 0.00804966, relative duality gap: 0.000200631
Solver 1 has terminated after 40 epochs in 15.8949 seconds
   Primal objective: 0.00487406, relative duality gap: 0.000584138
Time for the one-vs-all strategy
Time elapsed : 62.9996
\end{verbatim}
Note that the toolbox also provides the classes \texttt{LinearSVC} and \texttt{LogisticRegression} that are near-compatible with scikit-learn's API. More is available in the online documentation.

\section{Benchmarks}
We consider the problem of $\ell_2$-logistic regression for binary
classification, or multinomial logistic regression if multiple class are
present. We will present the results obtained by the solvers of Cyanure on 11
datasets, presented in Table~\ref{tab:datasets}
\begin{table}[hbtp!]
   \centering
\begin{tabular}{|c|c|c|c|c|c|}
   \hline
   Dataset & Sparse & $\sharp$ classes & $n$ & $p$ & Size (in Gb) \\
   \hline
   \texttt{covtype} & No & 1 & 581\,012 & 54 & 0.25 \\
   \hline
   \texttt{alpha} & No & 1 & 500\,000 & 500 & 2 \\
   \hline
   \texttt{real-sim} & No & 1 & 72\,309 & 20\,958 & 0.044 \\
   \hline
   \texttt{epsilon} & No & 1 & 250\,000 & 2\,000 & 4 \\
   \hline
   \texttt{ocr} & No & 1 & 2\,500\,000 & 1\,155 & 23.1 \\
   \hline
   \texttt{rcv1} & Yes & 1 & 781\,265 & 47\,152 & 0.95 \\
   \hline
   \texttt{webspam} & Yes & 1 & 250\,000 & 16\,609\,143  & 14.95  \\
   \hline
   \texttt{kddb} & Yes & 1 & 19\,264\,097 & 28\,875\,157 & 6.9 \\
   \hline
   \texttt{criteo} & Yes & 1 & 45\,840\,617 & 999\,999 & 21 \\
   \hline
   \texttt{ckn\_mnist} & No & 10 & 60000 & 2304 & 0.55 \\
   \hline
   \texttt{ckn\_svhn} & No & 10 & 604\,388 & 18\,432 & 89 \\
   \hline
\end{tabular} \caption{Datasets considered in the comparison. The 9 first ones can be found at \url{https://www.csie.ntu.edu.tw/~cjlin/libsvmtools/datasets/}. The last two were generated by using a convolutional kernel network~\cite{mairal2016end}. } \label{tab:datasets}
\end{table}

\paragraph{Experimental setup}
To select a reasonable regularization parameter $\lambda$ for each dataset, we first split each dataset into 80\% training and 20\% validation, and select the optimal parameter from a logarithmic grid $2^{-i}/n$ with $i=1,\ldots,16$ when evaluating trained model on the validation set. Then, we keep the optimal parameter $\lambda$, merge training and validation sets and report the objective function values in terms of CPU time for various solvers. The CPU time is reported when running the different methods on an Intel(R) Xeon(R) Gold 6130 CPU @ 2.10GHz with 128Gb of memory (in order to be able to handle the \texttt{ckn\_svhn} dataset), limiting the maximum number of cores to 8. Note that most solvers of Cyanure are sequential algorithms that do not exploit multi-core capabilities. Those are nevertheless exploited by the Intel MKL library that we use for dense matrices. Gains with multiple cores are mostly noticeable for the methods ista, fista, and qning-ista, which are able to exploit BLAS3 (matrix-matrix multiplication) instructions.
Experiments were conducted on Linux using the Anaconda Python 3.7 distribution and the solvers included in the comparison are those shipped with Scikit-learn 0.21.3.

In the evaluation, we include solvers that can be called from scikit-learn~\cite{scikit}, such as Liblinear~\cite{fan2008liblinear}, LBFGS~\cite{lbfgs}, newton-cg, or the SAGA~\cite{saga} implementation of scikit-learn. We run each solver with different tolerance parameter $\text{tol}=0.1,0.01,0.001,0.0001$ in order to obtain several points illustrating their accuracy-speed trade-off. Each method is run for at most 500 epochs.

\begin{figure}
   \subfloat[][Dataset covtype]{\includegraphics[width=0.32\linewidth]{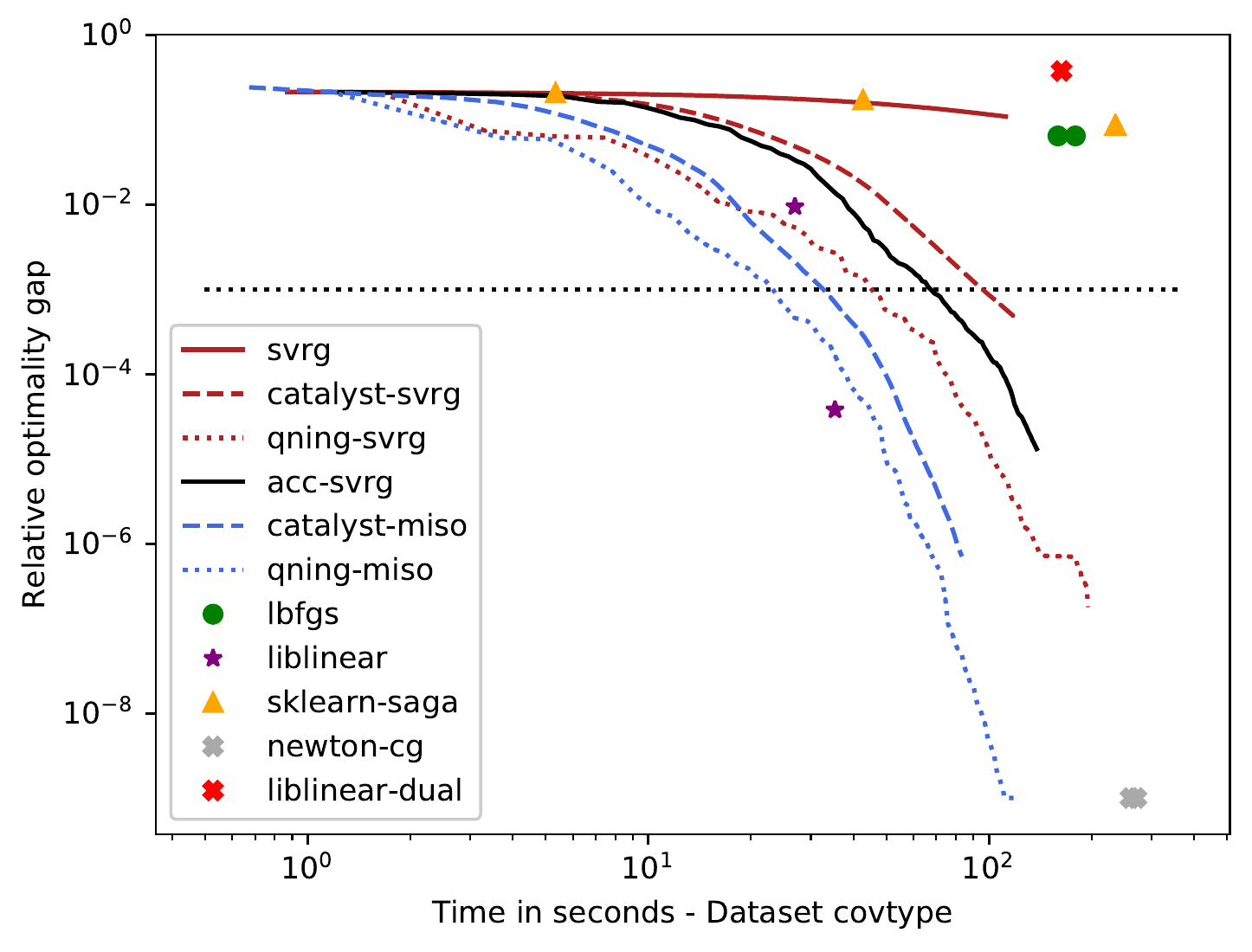}}
   \subfloat[][Dataset epsilon]{\includegraphics[width=0.32\linewidth]{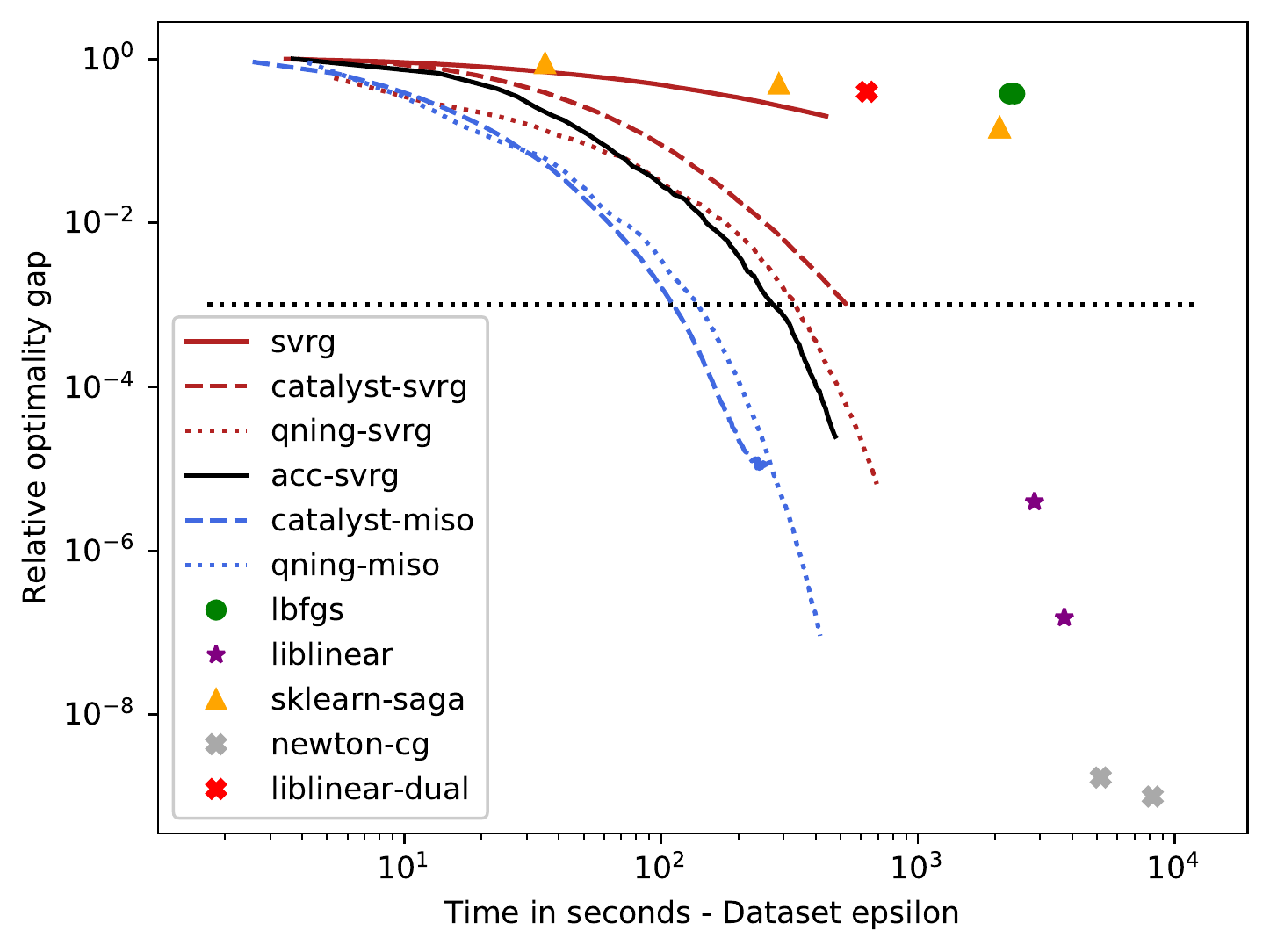}}
   \subfloat[][Dataset webspam]{\includegraphics[width=0.32\linewidth]{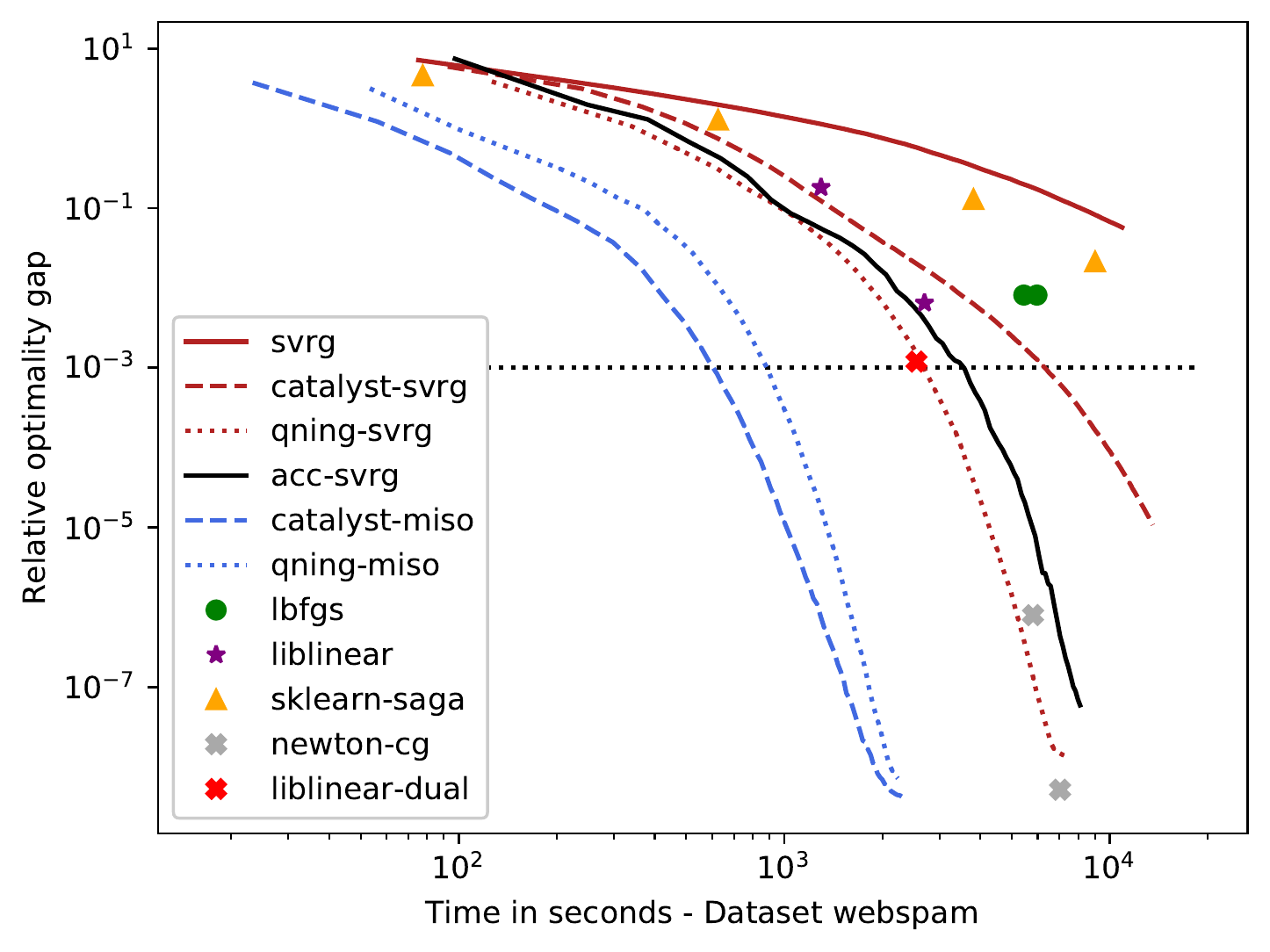}} \\
   \subfloat[][Dataset ckn\_mnist]{\includegraphics[width=0.32\linewidth]{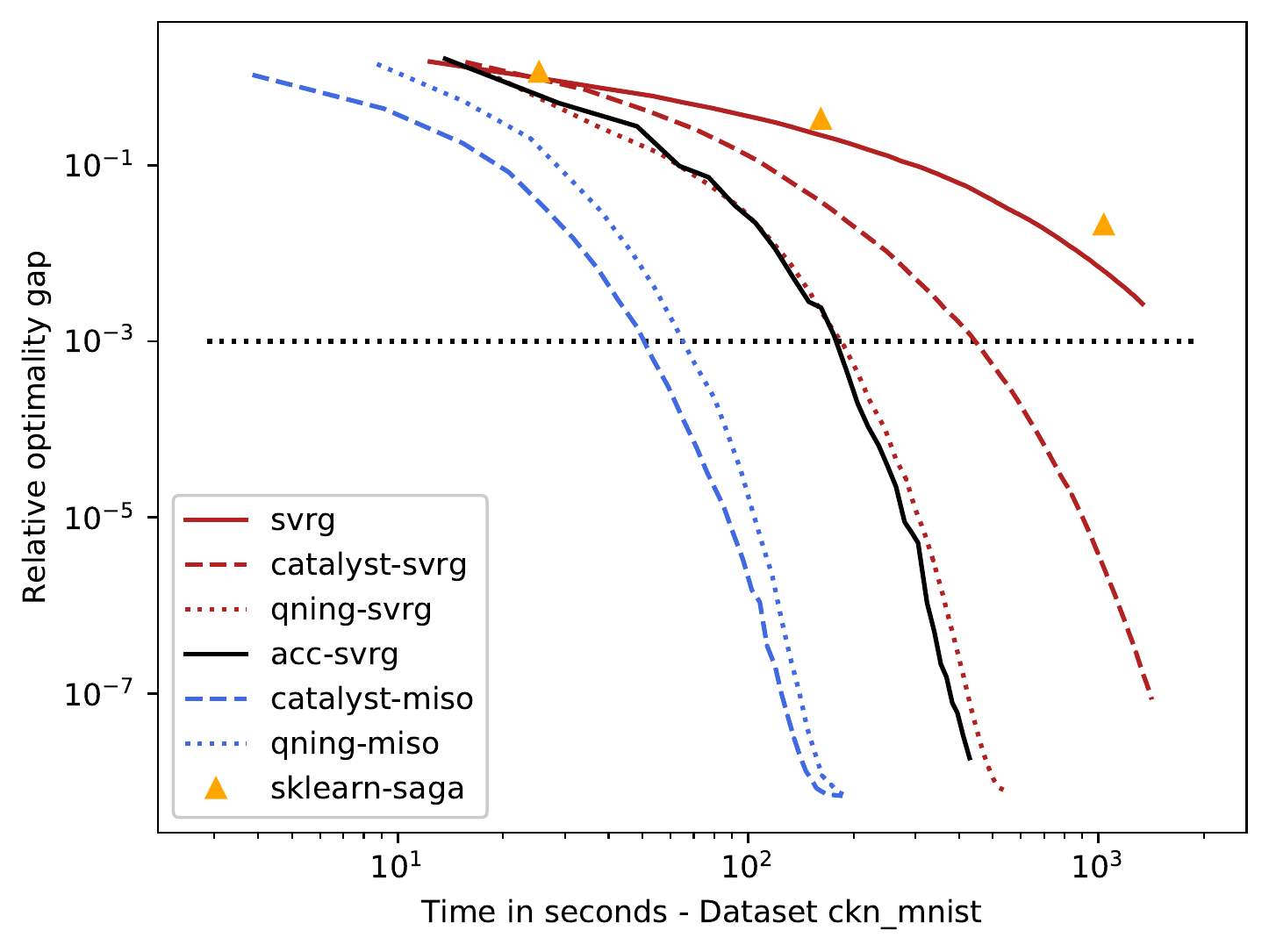}}
   \subfloat[][Dataset svhn]{\includegraphics[width=0.32\linewidth]{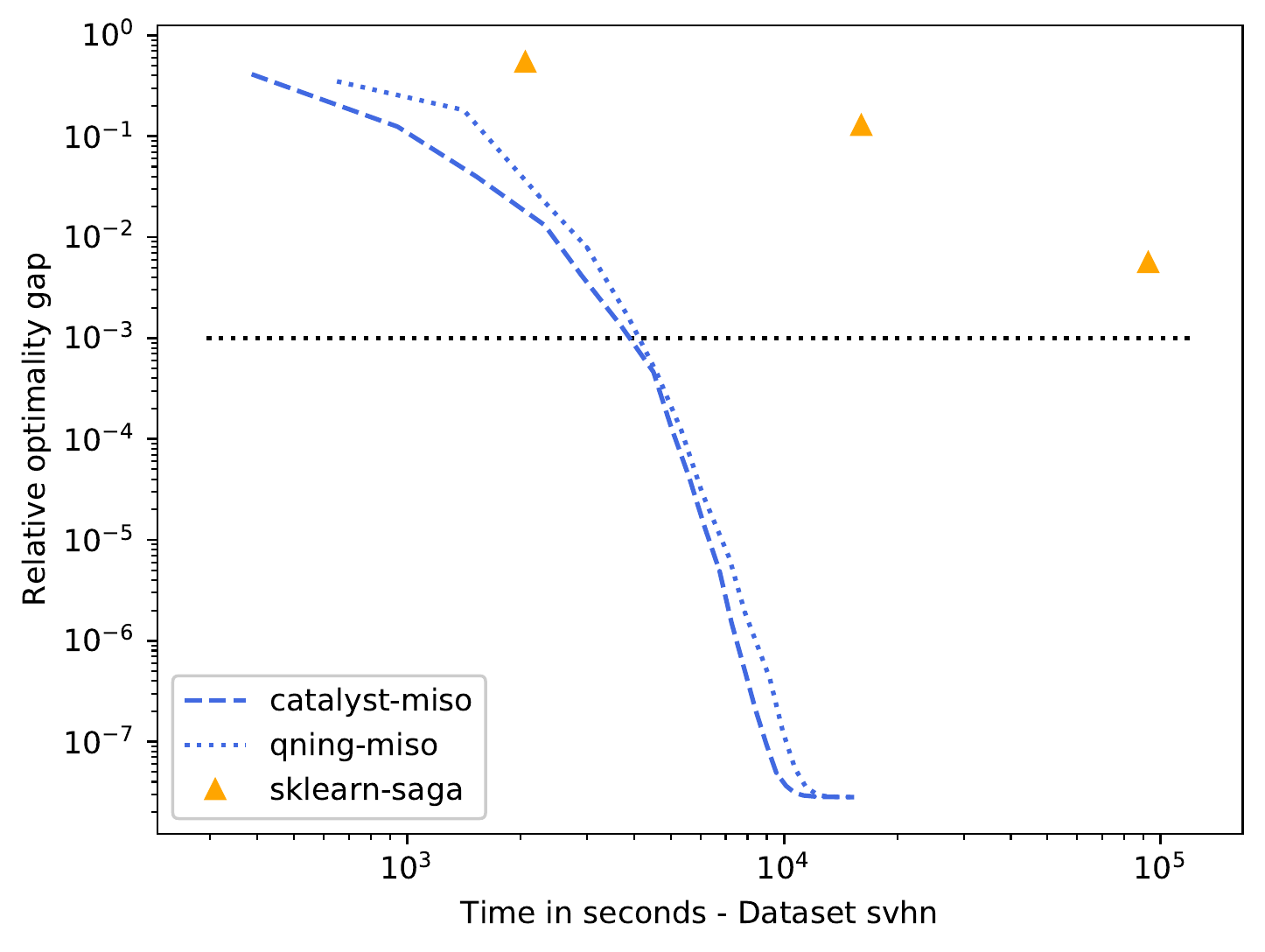}}
   \caption{Comparison for $\ell_2$-logistic regression for covtype, epsilon, webspam, ckn\_mnist, svhn -- the hard datasets.}\label{fig:bench1}
\end{figure}

\begin{figure}
   \subfloat[][Dataset alpha]{\includegraphics[width=0.32\linewidth]{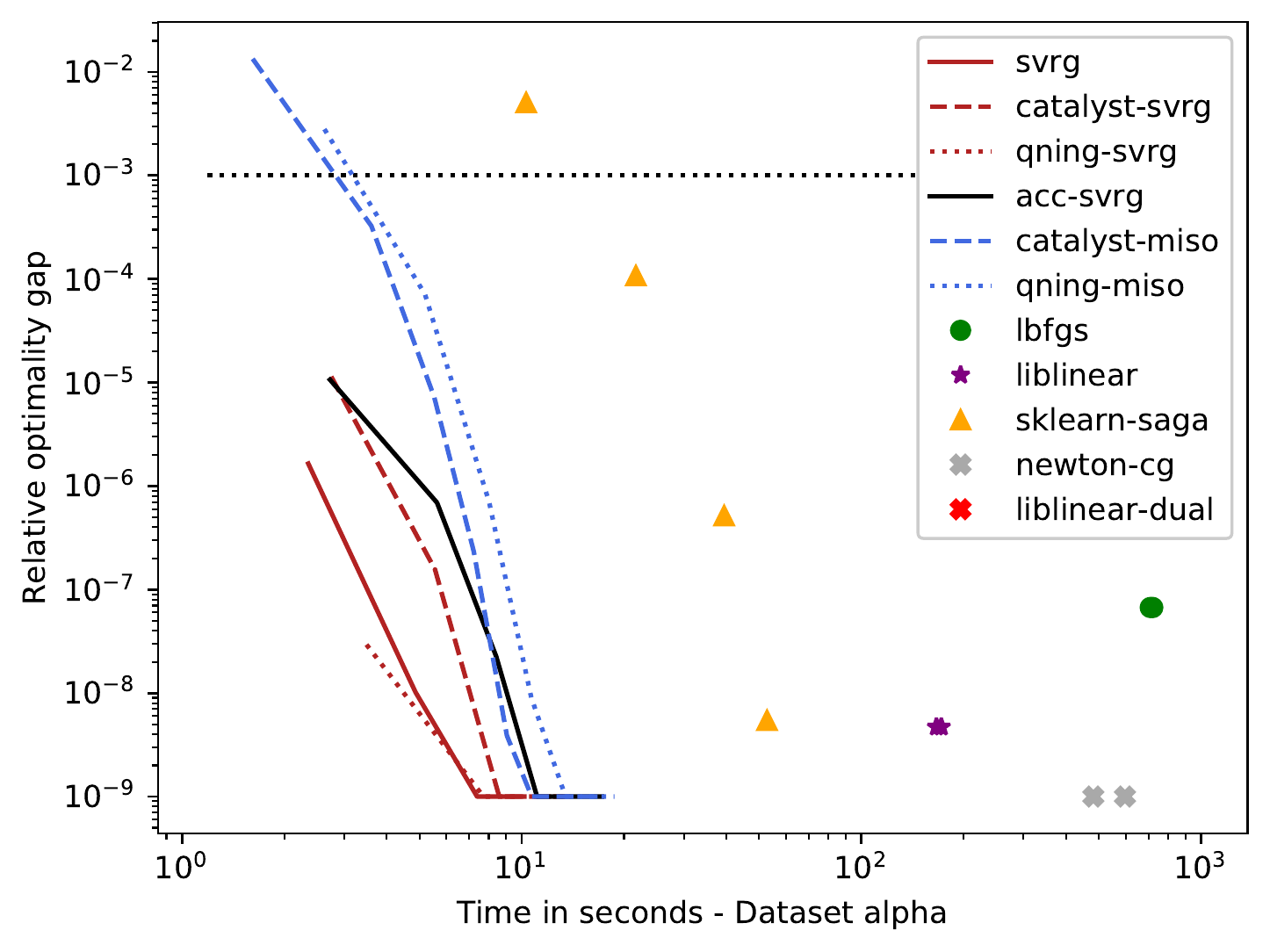}}
   \subfloat[][Dataset rcv1]{\includegraphics[width=0.32\linewidth]{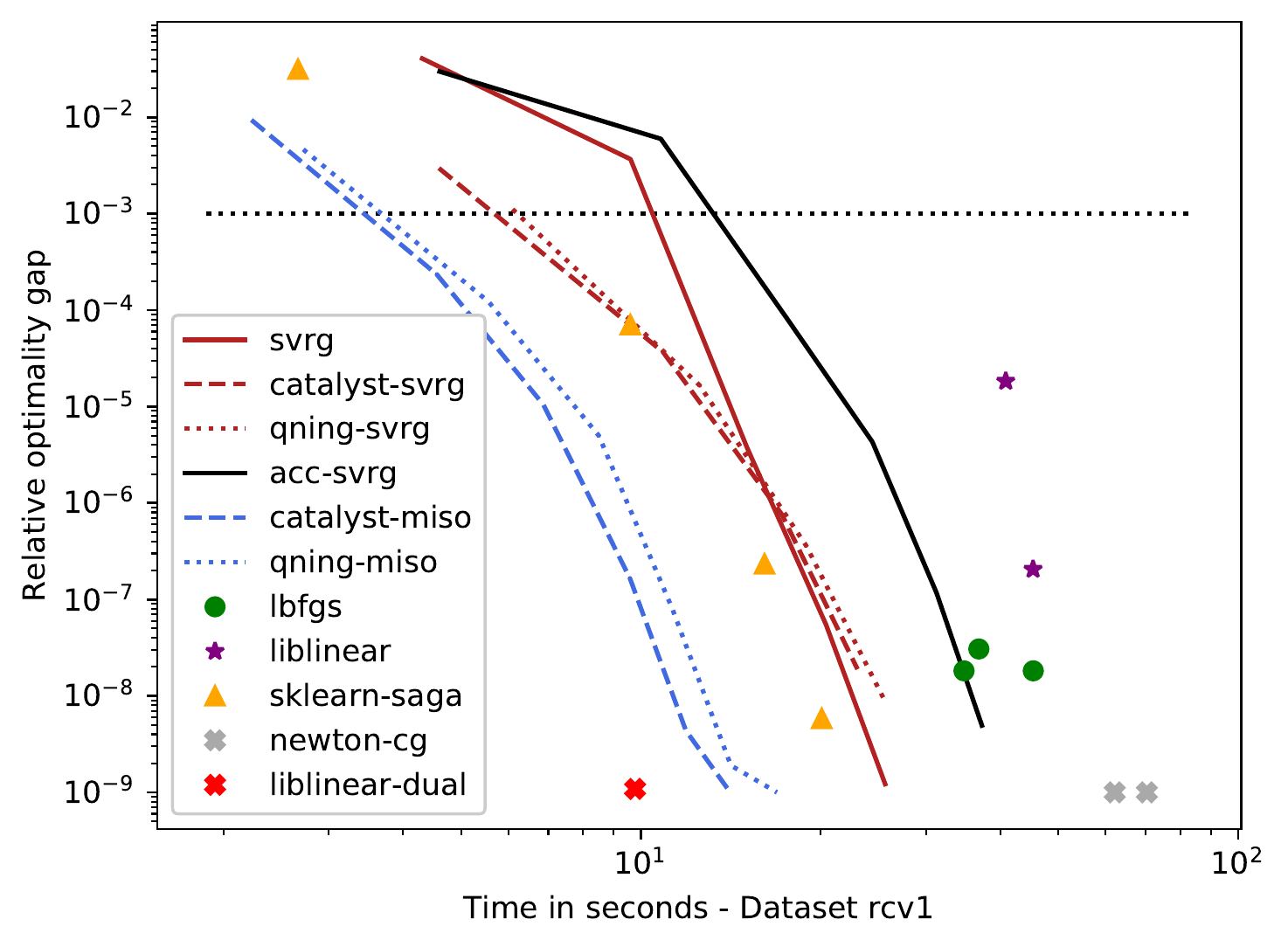}}
   \subfloat[][Dataset real-sim]{\includegraphics[width=0.32\linewidth]{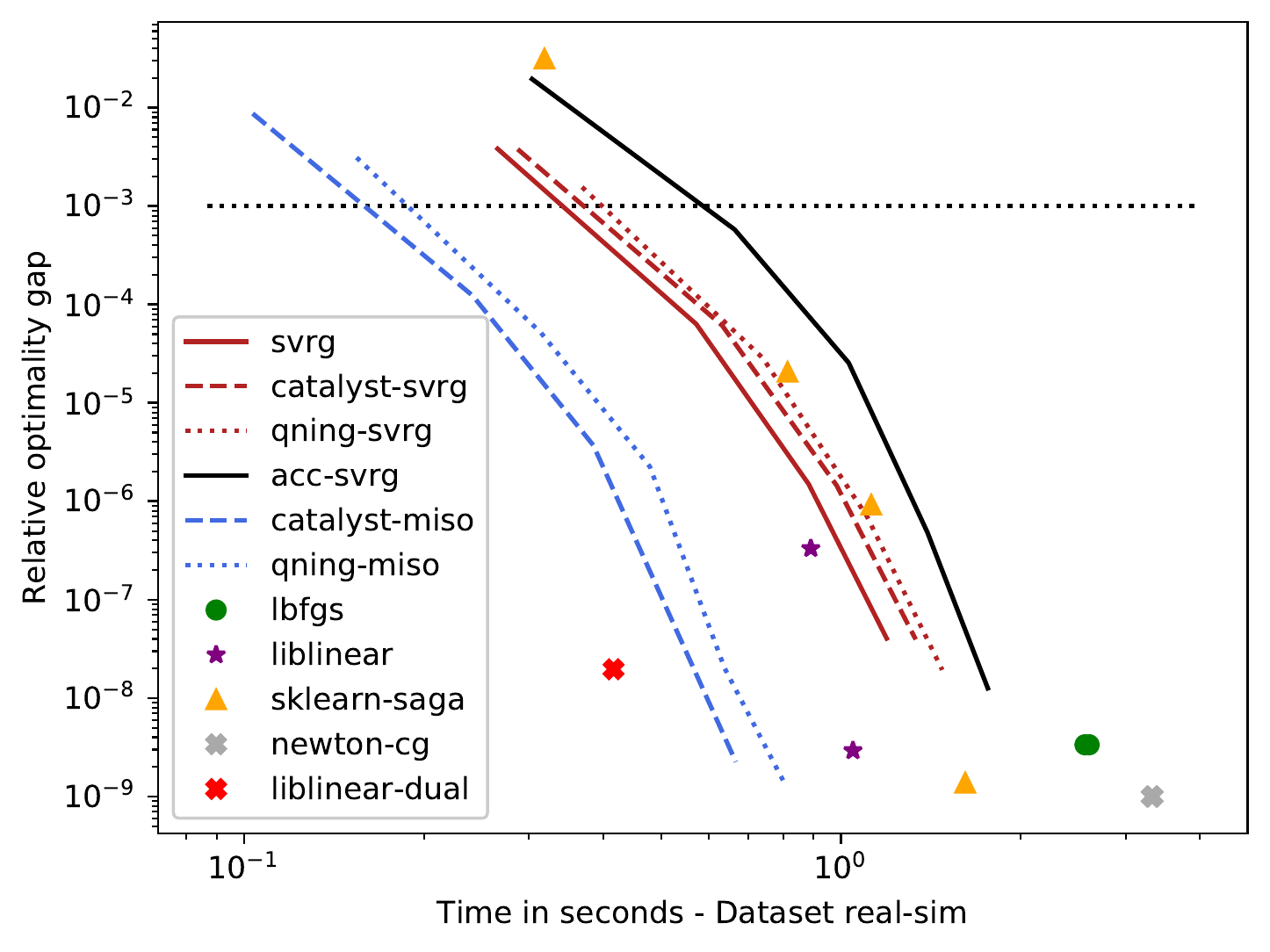}} \\
   \subfloat[][Dataset ocr]{\includegraphics[width=0.32\linewidth]{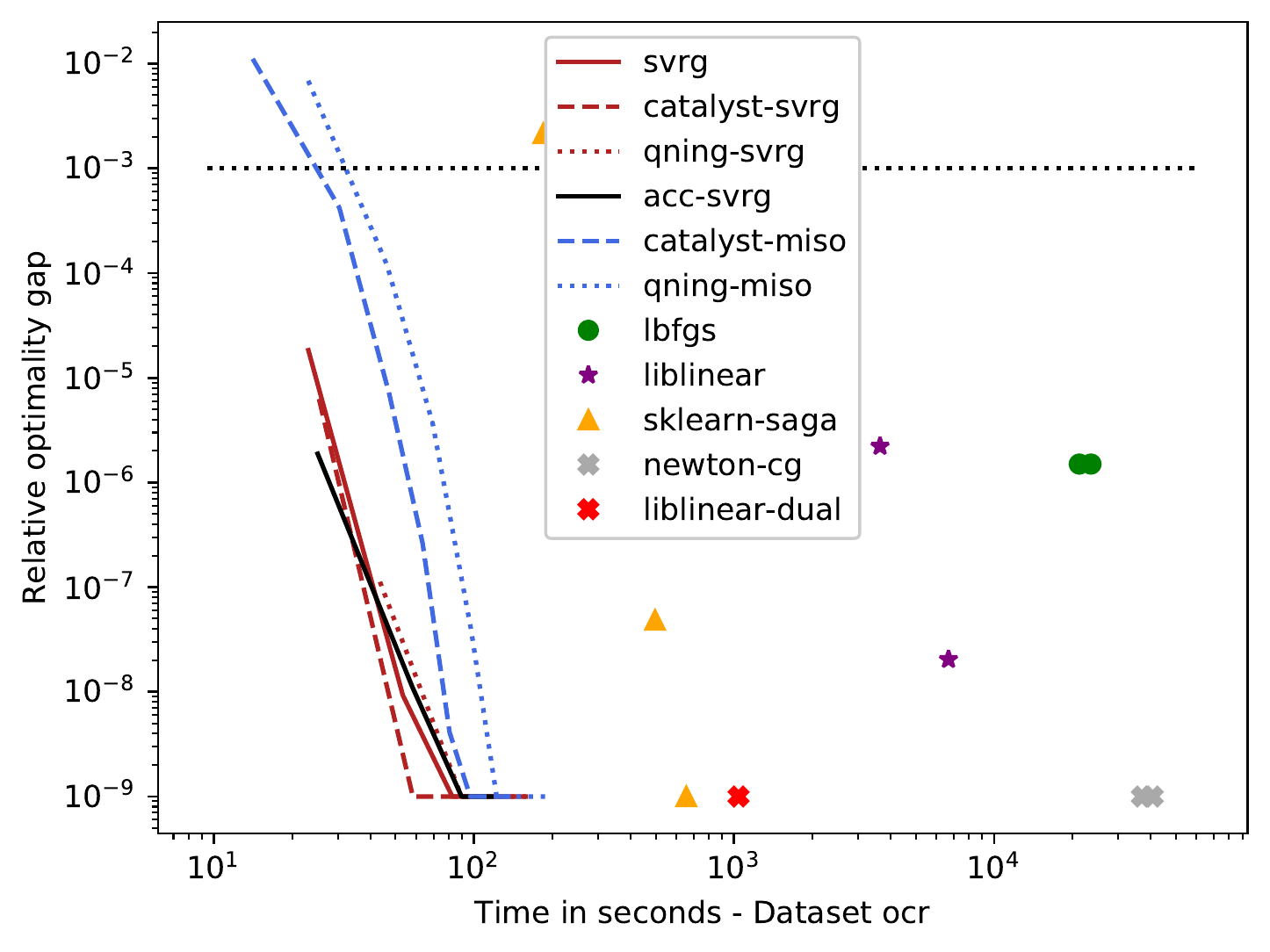}}
   \subfloat[][Dataset kddb]{\includegraphics[width=0.32\linewidth]{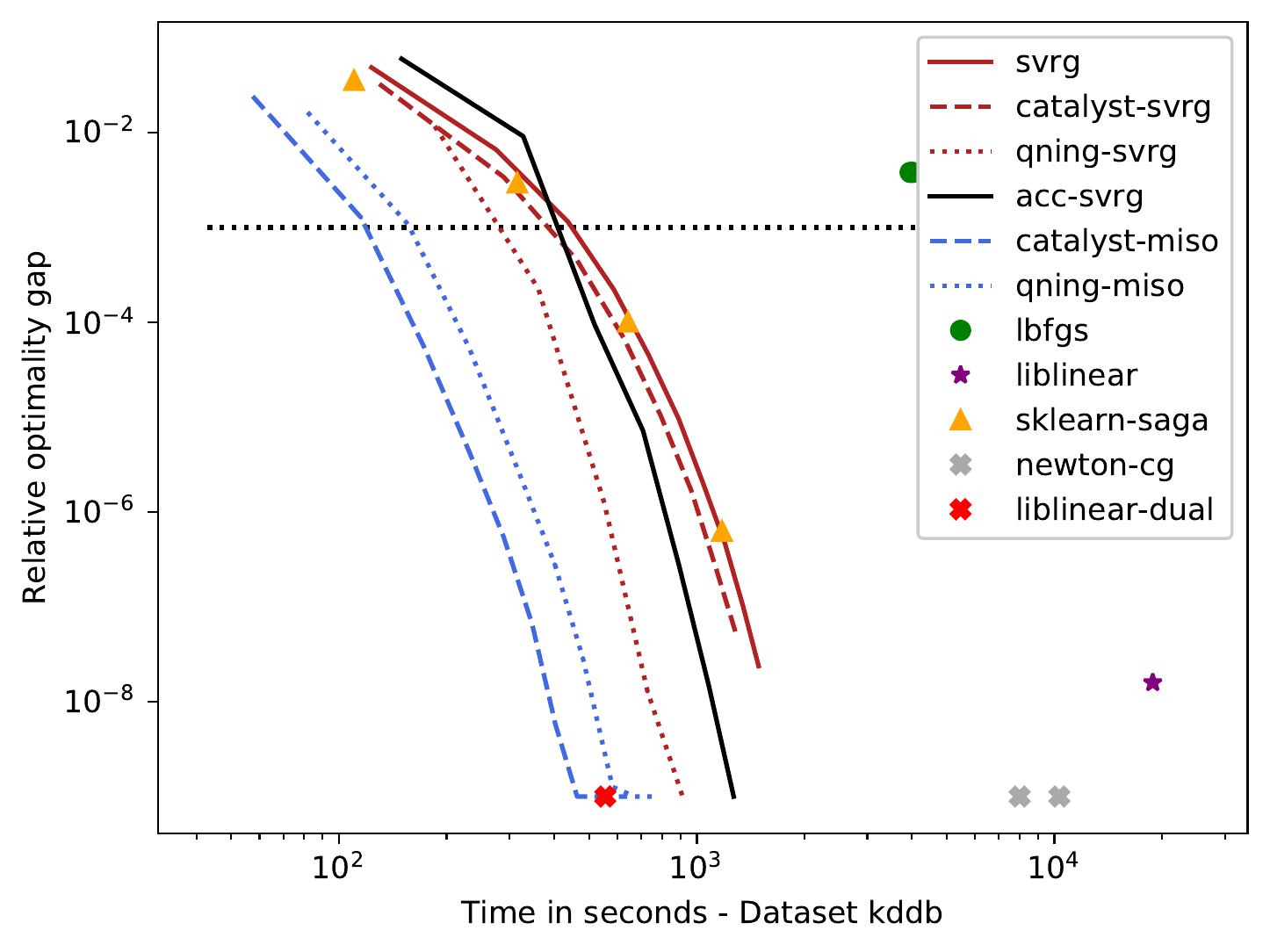}}
   \subfloat[][Dataset criteo]{\includegraphics[width=0.32\linewidth]{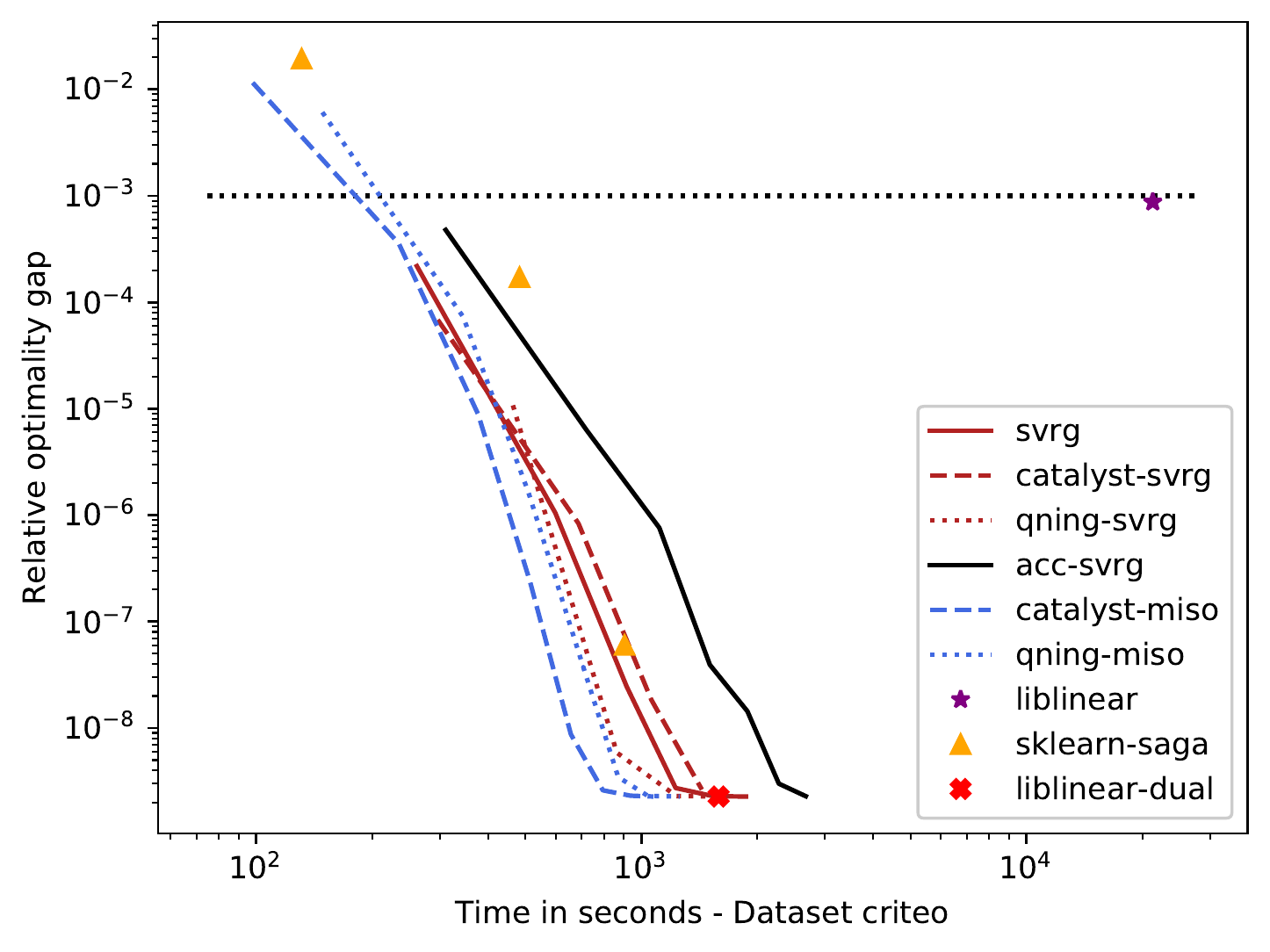}}
   \caption{Comparison for $\ell_2$-logistic regression for alpha, rcv1, real-sim, ocr, kddb, criteo -- the easy datasets.}\label{fig:bench2}
\end{figure}

There are 11 datasets, and we are going to group them into categories leading
to similar conclusions. We start with five datasets that require a small regularization
parameter (e.g., $\lambda=1/(100n)$, which lead to more difficult
optimization problems since there is less strong convexity. This first
group of results is presented in Figure~\ref{fig:bench1}, leading to the following
conclusions
\begin{itemize}
   \item \textbf{qning and catalyst accelerations are very useful}. Note that catalyst works well in practice both for svrg and miso (regular miso, not shown on the plots, is an order of magnitude slower than its accelerated variants).  
   \item \textbf{qning-miso and catalyst-miso are the best solvers here}, better than svrg variants. The main reason is the fact that for t iterations, svrg computes 3t gradients, vs. only t for the miso algorithms. miso also better handle sparse matrices (no need to code lazy update strategies, which can be painful to implement).
   \item \textbf{Cyanure does much better than sklearn-saga, liblinear, and lbfgs}, sometimes with several orders of magnitudes. Note that sklearn-saga does as bad as our regular srvg solver for these dataset, which confirms that the key to obtain faster results is acceleration. Note that Liblinear-dual is competitive on large sparse datasets (see next part of the benchmark).
   \item \textbf{Direct acceleration (acc-svrg) works a bit better than catalyst-svrg}: in fact acc-svrg is close to qning-svrg here.
\end{itemize}
Then, we present results on the six other datasets (the ``easy ones'' in terms of optimization) in Figure~\ref{fig:bench2}.
For these datasets, the optimal regularization parameter is close to $\frac{1}{n}$,
which is a regime where acceleration does not bring benefits in theory.
The results below are consistent with theory and we can draw the following conclusions: 
\begin{itemize}
   \item \textbf{accelerations is useless here, as predicted by theory}, which is why the 'auto' solver only uses acceleration when needed. 
   \item \textbf{qning-miso and catalyst-miso are still among the best solvers} here, but the difference with svrg is smaller. sklearn-saga is sometimes competitive, sometimes not.
   \item \textbf{Liblinear-dual is competitive on large sparse datasets.}
\end{itemize}

\subsection*{Acknowledgments}
This work was supported by the ERC grant SOLARIS (number 714381) and by ANR 3IA MIAI@Grenoble Alpes.

\bibliographystyle{plainnat}
\bibliography{bib}

\end{document}